# Space-contained conflict revision, for geographic information.

Omar Doukari[1], Robert Jeansoulin[2]

[1] LSIS, CNRS UMR6168, Centre Math Informatique, Joliot-Curie 39, 13453 Marseille Cedex 13, France. {omar_douk@yahoo.fr}

[2] IGM, CNRS UMR8049, Institut Gaspard Monge, Boul. Descartes, Champs sur Marne, 77454 Marne-la-Vallée Cedex, France. {robert.jeansoulin@univ-mlv.fr}

**Abstract.** Using qualitative reasoning with geographic information, contrarily, for instance, with robotics, looks not only fastidious (i.e.: encoding knowledge Propositional Logics PL), but appears to be computational complex, and not tractable at all, most of the time. However, knowledge fusion or revision, is a common operation performed when users merge several different data sets in a unique decision making process, without much support. Introducing logics would be a great improvement, and we propose in this paper, means for deciding –a priori- if one application can benefit from a complete revision, under only the assumption of a conjecture that we name the "containment conjecture", which limits the size of the minimal conflicts to revise. We demonstrate that this conjecture brings us the interesting computational property of performing a not-provable but global, revision, made of many local revisions, at a tractable size. We illustrate this approach on an application.

## Introduction

The purpose of this work is to address the difficulty of introducing logics in geographical knowledge processing, which is due to the intractability of the N-P-completeness of any algorithm that takes uncertainty into account.

What is the issue here? If data, involved in a logical formula, are uncertain and make this formula false, we name this a "conflict", and we can conclude that at least one of the data is of poor quality, and must be modified. But modified data can, in turn, yield new conflicts, and we must propose new modifications, and so on, following an exponential complexity. This fact has disqualified the classical logics from most of the real world applications, since long, and it has been illustrated several times with geographical data, where the combinatorial limit of such methods, does not pass a dozen of objects, with a few attributes, defined on domains of a small cardinality.

The idea is twofold: (1) though truth cannot be directly asserted, the *false* can be derived from inconsistency, at least what we may call a rational *false*, and (2) though consistency is only a global property of a whole set of knowledge, the union of several, locally consistent, and appropriately designed subsets, can be accepted as a non false solution. The gain that is expected, is not in terms of formal complexity, but simply in terms of cardinality. If the size of the subsets falls below the tractability limit (e.g. dozens of geographical objects), then we won, and we can iterate on the finite list of all subsets.

Main point is that we need to provide means to define what we name the "appropriate design", for the subsets of knowledge, which we will process independently. The idea for providing this, is based on the following empirical postulate: if a conflict is not detected, it can result from a lack of information, or, in case we have enough information, from an unreliable information; but, if we have enough, and reliable, information, then the absence of detected conflict can be interpreted as: no conflict at all. We name this idea, the postulate of « space containment » of information.

We briefly recall what is the belief revision and how it has been addressed by Reiter in his theory of diagnosis. We also present an illustration of cases where the user can make some guess on the appropriate design.

## Belief Revision

The general « revision problem » can be expressed as follows.

Let $S_1$ a finite and non empty set of beliefs (formulas), and $S_2$ another set of formulas (at least one), which are not beliefs, but reliable knowledge. Revising $S_1$ by $S_2$ means to make the union of all formulas, to check if they are altogether consistent, and, if necessary, to remove from $S_1$ (only) the minimal set of formulas whose removal restore the overall consistency, on remaining formulas.

This definition of the revision has received much attention since publication of the AGM axioms (Alchourrón 1985), and slightly modified versions (Gärdenfors 1988; Reiter 1987; Welkerson 1989; Papini 1992).

## Reiter's diagnosis approach of Revision (RDR)

The Reiter's algorithm for diagnosis (Reiter 1987), has since been applied and adapted by several authors. We have proposed (Würbel 2000), a version of this algorithm, adapted to the revision of $S_1$, a set of interval bounds (over a sampled domain), by $S_2$, a set of flux directions. Flux directions and interval bounds are related together by constraints, and interval bounds are said "in conflict", if the subset of constraints in which they are involved, cannot be satisfied (i.e. inconsistent). Let's recall what is the main notion behind this approach: the hitting-sets.

**Definition 1 (Minimal Hitting Set)**

Let $\mathcal{F}$ collection of sets. $F \subseteq \mathcal{F}$ is a hitting set iff: $F \subseteq \bigcup(G \in \mathcal{F})$, such that, $F \cap G \neq \emptyset$. F is a minimal hitting set of $\mathcal{F}$ iff: $\forall F'$, hitting set of $\mathcal{F}$, $F' \not\subset F$.

Let's note $\mathcal{N}(\mathcal{F})$ the set of all minimal hitting sets of $\mathcal{F}$. The Reiter's approach: (i) detects all inconsistent sub-sets of $(S_1 \cup S_2)$, into $\mathcal{I}(S_1 \cup S_2)$ ; (ii) computes $\mathcal{N}(\mathcal{I}(S_1))$ by removing all $e \in \mathcal{N}(\mathcal{I}(S_1 \cup S_2))$ that contain clauses from $S_2$; (iii) defines an order relation on $\mathcal{N}(\mathcal{I}(S_1))$, and chooses one preferred hitting set.

**Algorithm 1 (RDR,** *after Reiter' Diagnosis Revision***)**

```
I(S1∪S2): all inconsistent subsets of (S1∪S2).
T = ∅, root of the tree T.
If ∃s₀∈ (S1∪S2), then T = {s₀}.
Then ∀n, node of T,
   H(n) = {labelk | labels between n and root}.
   If n is labeled by Σ, in I(S1∪S2),
      ∀σ ∈ Σ, n has a successor on a branch labeled by σ
      if Σ∩H(n) = ∅.
   Else n has no successor in T, and we label n with ∅.
End.
```

**Property 1 (complexity of RDR)**

The maximal deepness of the tree $\mathcal{T}$ is $d=|S_1 \cup S_2|$. The formal complexity of the algorithm is in $O(d^3 \times 2^{2d})$.

Once all hitting sets are built, the most common approach consists in choosing one with minimal cardinality: this leads to "revise" the minimal number of original clauses. Depending on the formulation of the initial problem as a set of clauses (for instance propositional clauses), the resulting solution can differ, sometimes, from a solution that the expert would have qualified as "optimal". But we will not discuss this aspect here, and consider that we are never far from the expert optimal.

That global process is the regular way of using the diagnosis detection to perform revision.

## What's special with spatial Revision

The main and peculiar feature of "spatial information", is that any information is space related. With "geographical information", the Earth is the unique space reference. Hence, any proposition is necessarily confronted to a unique interpretation that we do not know, unfortunately. We only have beliefs, but they are not unique.

We call "space-domain" the basic representation of whatever is a "location" or a "space reference", and we call "region", or "parcel" a subset of the space domain. Any "item of geographic information" is either space-located by a direct mapping onto a set of parcels, or is derived by inference using one or several other items.

**Example 1 (simple spatial inference)**

Consider the statements: (1) there is a forest at parcel X; (2) all forests are green; (3) roads that cross a forest are dark.
A very simple representation in PL, the propositional language, can be:
- To list parcel properties (finite): such as Fx (forest at x) or ¬Fx (no forest),
- To infer consequences from rules (2) and (3), on this list: Fx $\rightarrow$ Gx (green) and Fx $\wedge$ Rx $\rightarrow$ Dx (dark).

"Who acts like this?", Nobody! Or may be everybody, at some stage of the design of the application. We can imagine situations where there are great numbers of parcels to process, with constraints that we cannot check by visual inspection, and when several possible, but incompatible, corrections would probably improve the overall analysis. It is likely to consider that space constraints are a strong basis for such situations to occur.

This is what we address here, and therefore it is important to question the "tractability" of such an approach.

## Space Containment

The idea of "space containment" is based on this empirical postulate: if a conflict is not detected, this may be because of a lack of information, or, in case we have enough information, because of some unreliable information; but, if we have enough information, and reliable information, then the absence of detected conflict can be interpreted as if there is no conflict at all.

## Ontology for Space containments

In order to formalize this postulate, we need to agree on a minimal ontology for representing space. For our goal, we do not need much geometry for representing the geographic space. We need a basic topology for the space-domain, because we want to allow distinction between the inside and the outside of a given region.

Let's use some simple and graph-theoretic definitions (Weisstein 1999).

### Definition 2 (Simple Planar Graph)

A graph is planar if it can be drawn in a plane without graph edges crossing. A simple graph, is an un-weighted, undirected graph containing no graph loops or multiple edges.

### Definition 3 (Path)

A *path* is a non-empty graph $P = (V, E)$ of the form $V = \{v_0, v_1 ..., v_k\}$ and $E = \{v_0v_1, v_1v_2 ..., v_{k-1}v_k\}$, where all $x_i$ are distinct. The vertices $v_0$ and $v_k$ are linked by $P$ and are called its *ends*; the vertices $x_1,...,x_{k-1}$ are the *inner* vertices of $P$. The number of edges of a path is its *length*.

### Definition 4 (Distance)

Between two vertices of the graph, we define the distance as the length of the minimal path between them. Notation $d(v_1, v_2)$.

### Definition 5 (k-Neighborhood)

The k-Neighborhood of a vertex $v_0$, is the set of all vertices $v_i$, such that $d(v_0, v_i) \leq k$.

**Definition 6 (q-Containment)**

Given a k-Neighborhood $N_0$ of a vertex $v_0$, and a k'-Neighborhood $N'_0$ of the same vertex $v_0$, with k'>k, the set difference $N'_0 - N_0$ is called a q-Containment, and q= k'-k is the thickness of the containment.

**Example 2 (trivial topology on a space partition)**

Splitting up a country into "counties", leads to a partition, where every space location falls into one and only one county. The trivial topological space is the power set of the set of counties. The empty set $\varnothing$ and the country itself (union of all counties) are members of this topology.

## Splitting up Space into containments

We propose an algorithm *Partition*($V$={vertex_set}, k, k'), for decomposing the space into q-Containments, according to two integers, k, k'.

This algorithm starts from a "seed vertex" $v_0$, randomly chosen in the set $V$ of all vertices. Thus, this algorithm is parameter dependant: we will discuss this point later. Then, all vertices in the k-Neighborhood of vertex $v_0$, form the first block $B_0$, of the partition, and are removed from $V$. A second seed is chosen, and so on, until $V$ is empty.

**Notation 1 (Blocks and Covers)**

Space $S = \{v_0, v_1 \ldots v_m\}$ is a set of vertices (i.e.: nodes of a graph of parcels).

$B_\alpha(S) = \{B_0, B_1 \ldots B_n\}$ is one partition of $S$ in k-Neighborhoods.

$Q_\alpha(S) = \{Q_0, Q_1 \ldots Q_n\}$ is the covering in q-Containments associated to $B_\alpha(S)$. A *block* is an element of $B_\alpha(S)$. A *cover* is its corresponding element in $Q_\alpha(S)$.

At this stage, the "parcels" are vertices, and we can index all information, with reference to blocks and covers, which contain the parcels. For sake of simplicity and generality, we propose to represent knowledge by clauses in Propositional Logics (PL). The terms "parcel" and "vertex" are used as equivalent, as well as "proposition" and "clause".

**Notation 2 (Clause "Belongs-to-block")**

If a proposition denotes a property for a vertex $v_x$, noted $Pv_x$, then we say that this proposition "belongs-to-" the block of $B_\alpha(S)$, which contains $v_x$.

For a composition of clauses, e.g.: $C = (Pv_x \vee P'v_y)$, we say: $C$ belongs-to- $B$, the <u>smallest</u> k-Neighborhood containing both $v_x$ and $v_y$. $B$ is a k-Neighborhood (union of two) in one of four situations:

1. $\exists B_1 \in B_\alpha(S), B \subseteq B_1$
2. $\exists Q_1 \in Q_\alpha(S), B \subseteq (Q_1 \cup B_1)$, $Pv_x$ belongs-to $B_1$ and $P'v_y$, belongs-to $Q_1$

3. $\exists Q_1 \in Q_\alpha(S)\ B \subseteq Q_1$, that is: neither $Pv_x$ nor $Pv_y$ belongs-to $B_1$
4. *None of the above.*

**Notation 3 (b-clause, c-clause, q-clause, and n-clause)**

We name *b-clause* a clause being in situation 1 (fully in a block)

*c-clause* a clause crossing between a block and its cover,

*q-clause* a clause fully and strictly in a cover (not in a block),

*n-clause* a clause in any other situation.

**Example 3 (2D-space partition)**

The figure illustrates a space partition, with some adjacency links (edges), and a chain (bold edges). The distance is symbolized by decreasing darkness.

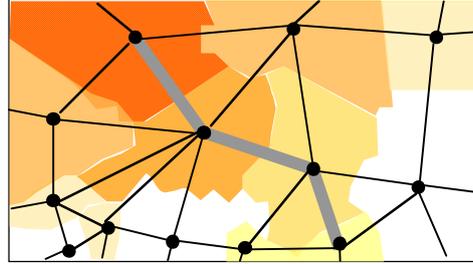

**Figure 1: Adjacency (graph), Chain (bold), and Distance (# of nodes).**

## Defining conflicts, consistency and the containment conjecture

We need some definitions about conflicts. Let $C = \{c_0, c_1 \ldots c_m\}$ the set of all clauses of a problem.

**Definition 7 (Minimal conflict)**

A conflict is an inconsistent subset of clauses (according to the chosen logics, say PL). A minimal conflict $C$ is a conflict such that: $\forall C' \subset C$, $C'$ is consistent.

**Definition 8 (Size of a conflict)**

The size of a conflict is $k_0$ the size of the smallest k-Neighborhood to which belong all the clauses of the conflict.

Now we can formulate the "*containment conjecture*".
Let $B_\alpha(S) = \{B_0, \ldots B_n\}$ a partition of $S$ in k-Neighborhoods. The conjecture says:

$\forall C' \subseteq C$, such that $\exists B_i \in B_\alpha(S)$ and $C' = \{clauses\ of\ B_i\}$,
- if $C'$ is consistent, and

> - if $C' \cup \{clauses\ of\ Q_i\}$ is consistent,
> then: $C'$ is consistent in $C$.

[i.e.: if every restriction of the knowledge base to a block is independently consistent, and if every such restriction is consistent with its q-Containment, then this restriction is conflict-free with any other information.]

A direct application of such a conjecture is that it would be sufficient to perform n partial revision of the entire knowledge base, then to check these partial models within their respective containments, and –if everything is all right- to accept as a global model the simple union of all the partial models. The case where something goes wrong is discussed in a next section. In conclusion, knowing that revision is a NP-complete process means that there is a cardinality beyond which the computation becomes intractable. Therefore, restricting the computation to subsets of a smaller cardinality, can provide a solution for the problem.

*Remark 1 (**Para-consistency**). Obviously, there is no guarantee that the "containment conjecture" is true, and, precisely because we want to use it for saving to check the full consistency, we don't want to verify it. This leads to a para-consistent situation, that is, we accept inconsistency as long as not all inferences will be performed during the application, see (Mares 2002). Saying it differently: the conjecture says that some consequences of our belief will never be examined. In real life, this kind of assumption is used almost everyday, when we say that something will never happen.*

## Processing Space contained Conflicts

Our purpose now, is to demonstrate that the "containment conjecture" can be replaced by a much simpler condition, which involves only the size of the minimal conflicts. We name this condition the *hypothesis H0*. It relates the size of the containments with the bound on tractability.

**Definition 9 (Hypothesis H0)**

> If $k_t$ is the size of the largest tractable k-Neighborhood, if $d_c$ is the maximal size of the minimal conflicts, then the problem is tractable if: $d_c \leq k_t/3$.

Three types of conflicts can occur, depending on the type of clauses involved in the conflicts. This section doesn't detail demonstrations, they can be found in (Doukari, Msc. thesis).

### Space Independent Conflicts.

This is the case when all conflicts do not contain any *c-clauses*. Hence, if we meet a conflict, it is only composed of *b-clauses*, and whence we have performed a revision for in $B_i$, it will hold also for $Q_i$. In other words, from a computing viewpoint, the search for the minimal hitting-sets of the subset of clauses belonging to the block (*b-clauses* of $B_i$) can ignore the conflicts in the other -non overlapping- blocks.

A special version of the algorithm Reiter'Diagnosis RDR, has been designed for this purpose. It has the same theoretical complexity, but it processes less clauses. And, as a noticeable consequence, we can remove the block from the list of the blocks to process, because it will not interfere anymore. We do not detail this here: the general version encompasses it.

### Information Independent Conflicts.

In this case, some conflicts can contain *c-clauses*, but their intersection with others conflicts containing only *b-clauses*, should be empty. Again, the search for minimal hitting sets can be made independently of other blocks, but we cannot remove the block from the list, because some conflicts are still depending on it. In order to process these conflicts, we need to shift the blocks for finding all blocks containing independent conflicts. When shifted blocks (typically one or two) are processed, the other minimal hitting sets can be added to the first list, and the augmented list constitutes exactly the hitting set of the union of all the conflicts involved. This result has been established formally (demonstration omitted).

**Proposition 1**

If two sets of clauses are independent, the minimal hitting sets of their union are the set of the union of their local minimal hitting sets.

The important consequence, for our purpose, is that the extra cost between this case and the previous one, is simply polynomial. The complex part (Reiter) is computed with the same basic cardinality k, size of the blocks, and not k', size of the block plus its cover. The total number of shifts is *ns* = $2\times(k'-k)$ in the one dimension case (the example), and *ns* = $4\times(k'-k)^2$ for geographical information.

**Dependent Conflicts.**

First we process, block-by-block the conflicts that contain at least one *b-clause*, then we can ignore conflicts already processed or conflicts containing only *q-clauses*, because they will be considered later in one of the next block (because of the conjecture). Hence, for each block, the minimal hitting sets will be made only of conflicts that contains only *b-clauses* and conflicts that contains at least one *c-clause*. Each of these minimal hitting sets is concatenated with previously computed hitting sets, and we keep the minimal ones. The clauses of the block are removed from the list, hence the computation will become faster and faster.

**Proposition 2**

Let $\mathcal{F} = \{f_0, f_1, \ldots f_p\}$ a collection of set of clauses: $\mathcal{F} = \cup_{k=1,p} f_k$, $f_k \cap f_l = \emptyset$ if $k \neq l$. If $H_0, H_1, \ldots$ are the respective minimal hitting sets of $f_0, f_1, \ldots f_p$, then $\mathcal{H} = \text{Min}(\{\cup_{k=1,p} h_k\} | h_k \in H_k\})$ is the set of global minimal hitting sets for $\mathcal{F}$.

The overall process is named "*Contained-Revision*", to recall that the containment conjecture is necessary, and that, without it, the result is not guaranteed to be a model (i.e.: consistency is not proved, since conjecture is not proved, but only accepted as true).

**A geographical example of belief revision.**

Before presenting the algorithm for Contained-Revision, let's consider a more thorough example, which illustrates various aspects of our approach.

*Example 4: representation and processing of flood data*
In this real example, data are incompletely known, and rather approximate, for about 60% of the parcels. Fluxes are incompletely known, and only the direction is visible (no quantitative assessment). A qualitative approach of the problem seems reasonable, if not effective.
Figure 2 shows parcels and explains how "water level interval" (data source S1) and "flux" (data source S2) are related in the representation of a flooding.

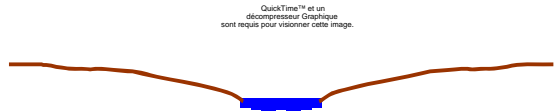

**Figure 2: an interval-based representation of a flooding.**

In Figure 3, we have two models, built on blocks of 4, and 3 elements respectively. We notice that a local consistency check is not sufficient for detecting some conflicts that are "hidden". We need to widen the check to an appropriate "cover".

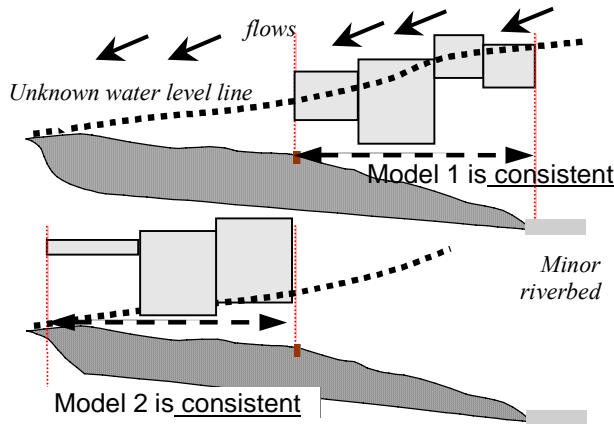

**Figure 3: two partial models, independently consistent.**

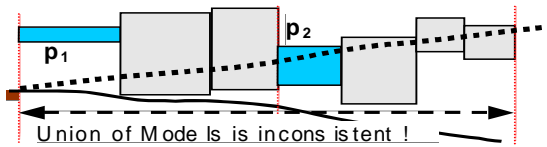

**Figure 4: the union of two consistent models can be inconsistent.**

Figure 4 shows a conflict between $p_1$ and $p_2$, which contains a c-clause, for k1=3 and k2=4. The conjecture is that the size of "hidden" conflicts is bound by a limit, and that this limit is 4 in this example.

Figure 5 shows how to shift blocks and covers in order to capture the missed conflicts containing *c-clauses*: '*shift_1*' misses the conflict because $p_1$ and $p_2$ are too close to the center of their respective blocks $B_2$ and $B_3$, but '*shift_2*' allows to view $p_1$ and $p_2$ in the same cover $Q'_2$.

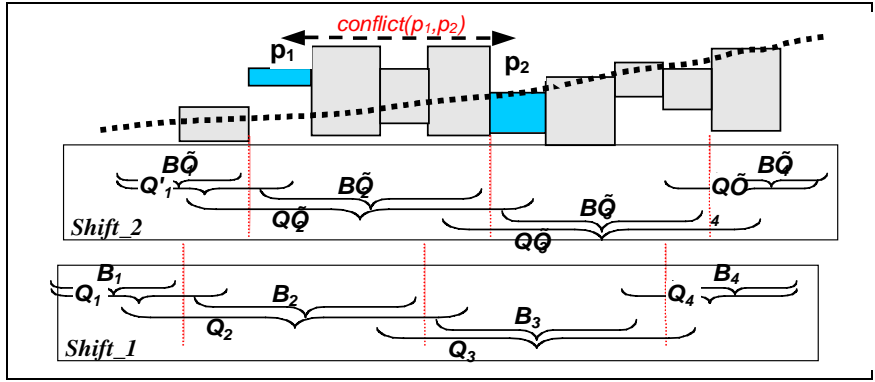

**Figure 5: shifts capturing or not capturing conflicts.**

## Containment-based Revision.

The full algorithm for *Contained-Revision* uses the routine *Partition*, and the modified *RDR*, as previously defined, and implements the whole processing.

**Algorithm 2 (*Contained-Revision*)**      [simplified version]

```
Hglobal= Ø. Conf= Ø (revised conflicts).
Blocks = Partition(S, k, k');
[a first step for filtering independent conflicts can be added for
improving the efficiency: omitted here]
for all Bi in Blocks;
   Htemp = Ø;
   if Hi = RDRspecial(union(Bi,Qi) - Conf) • Ø;
         for all hitting set hk in Hglobal;
            to update Htemp with hk and Hi;
         end forall
      to update Htemp if not minimal;
      to update Hglobal;
      to increment Conf;
   endif
   to decrement Blocks;
end forall
return Hglobal.
End.
```

Routine *union(bi,qi)* performs the mere set-union of the block with its cover.

**Proposition 3 (Cardinality Reduction)**

The practical complexity of the Contained-Revision algorithm reduces the complexity of the original RDR, by a polynomial factor.

The complexity of RDR has been given: $O(d^3 \times 2^{2d})$, with $d = |S_1 \cup S_2|$. With the same d, and with m = Cardinal(S), the *Contained-Revision* complexity is in $O(m \times d \times 2^{2d})$. From a pure formal viewpoint, only the exponential term is important, because it determines the intractability alone. But it is worth noticing that the ratio is polynomial, $r = (d^2 / m)$.

Let's consider the real scale "flooding" application: 300 parcels, 2 attributes per parcel, defined over a sampled domain of water heights (about 10-20 useful sample level by parcel). The representation of this problems leads to about 100.000 propositional clauses, hence $d = 10^5$, and $r = (10^{10} / 300) = 3.10^7$. It is 30 million times faster. Though it is useless when we pass the tractability threshold, it would be worth to save this amount of time when we are near the threshold.

## Conclusion.

We have found a condition, that we name the "*containment conjecture*", which allows to use a revision operator, similar to the Reiter' diagnosis algorithm, under the assumption that $d_c$ the maximal size of minimal conflicts, be smaller than $k_t$ the maximal size of tractable k-Neighborhood.

If it is rather simple to guess what can be $k_t$, it is not always possible to make an assumption on the maximal size of minimal conflict. At least, we may ask: Can the application reasonably avoid checking the consistency for data that are apart enough from each other? And by a distance such that, if no contradiction has been detected (in between), then it means that there must not be contradiction at all?

Something like: if my traffic data are consistent with my expectation, and are not in conflict with data in the surrounding vicinity, why should I check an improbable conflict with data much far away.

**Acknowledgments.** This work is supported by *Region Provence-Alpes-Cote-d'Azur* which provides a doctoral grant for M. Doukari, and is directly continued from the REV!GIS European project IST-1999-14189.